\documentclass[conference]{IEEEtran}
\IEEEoverridecommandlockouts
\usepackage{cite}
\usepackage{amsmath,amssymb,amsfonts}
\usepackage{tabularx}
\usepackage{algorithmic}
\usepackage{booktabs}
\usepackage{enumitem}
\usepackage{subcaption}
\usepackage{graphicx}
\usepackage{textcomp}
\usepackage{xcolor}
\def\BibTeX{{\rm B\kern-.05em{\sc i\kern-.025em b}\kern-.08em
    T\kern-.1667em\lower.7ex\hbox{E}\kern-.125emX}}
\begin{document}

\title{Embodied Referring Expression for Manipulation Question Answering in Interactive Environment\\
\thanks{This work was supported by the Seed Fund of Tsinghua University (Department of Computer Science and Technology)-Siemens Ltd., China Joint Research Center for Industrial Intelligence and Internet of Things.}
\thanks{$^{\dagger}$Corresponding author:Huaping Liu (\texttt{hpliu@mail.tsinghua.edu.cn})}
}

\author{\IEEEauthorblockN{Qie Sima$^{1}$, Sinan Tan$^{1}$, Huaping Liu$^{1,2,\dagger}$}
\IEEEauthorblockA{$^1$Department of Computer Science and Technology, Tsinghua University}
\IEEEauthorblockA{$^2$Beijing National Research Center for Information Science and Technology, China}
}

\maketitle

\begin{abstract}
Embodied agents are expected to perform more complicated tasks in an interactive environment, with the progress of Embodied AI in recent years. Existing embodied tasks including Embodied Referring Expression (ERE) and other QA-form tasks mainly focuses on interaction in term of linguistic instruction. Therefore, enabling the agent to manipulate objects in the environment for exploration actively has become a challenging problem for the community. To solve this problem, We introduce a new embodied task: Remote Embodied Manipulation Question Answering (REMQA) to combine ERE with manipulation tasks. In the REMQA task, the agent needs to navigate to a remote position and perform manipulation with the target object to answer the question. We build a benchmark dataset for the REMQA task in the AI2-THOR simulator. To this end, a framework with 3D semantic reconstruction and modular network paradigms is proposed. The evaluation of the proposed framework on the REMQA dataset is presented to validate its effectiveness.
\end{abstract}


\section{Introduction}
Recently, the AI community has witnessed the prosperity of Embodied AI where agents are required to perform tasks in various forms with egocentric vision. The success of embodied AI brings up the interest of researchers in the robot community to transfer methods in off-shelf Embodied AI tasks to robot platforms.

Currently, most of works in Embodied AI have revolved around the task of navigation – including position-goal, object-goal, and area-goal\cite{duan2022embodiedAIsurvey}. However, the ability to actively manipulate objects and physically interact with the environment becomes crucial in the embodied robot task, where agents need to perform complex tasks in the real world. As the studies on embodied tasks have surged in recent years, a wide variety of embodied tasks has been proposed. However, very few works have looked into a general framework for embodied tasks that involve most modular models of robot task in the real world: Visual reception, Language comprehension, Active navigation and Manipulation. In an embodied robot task, how to localize the target object precisely and effectively has always been a challenge. Since many objects in the real scene are similar in shape and appearance (e.g., books on shelf, cabinets in the kitchen).

Referring Expression (RE) is a widely studied cross-modal task in both computer vision and natural language processing fields as a vision and language task. In a RE task, the agent needs to localize a specific target object in the image in response to a given natural language referring expression. Most of current studies in referring expression focus on passive image datasets (e.g. RefCOCO, RefCOCO+\cite{2014ReferItGame}, RefCOCOg\cite{yu2016modeling}) where samples will not change with agent's decision. Recently, referring expression tasks in embodied scenarios has emerged. In an Embodied Referring Expression (ERE) task, the agent is required to navigate to the position mentioned in the given expression in a 3D environment and complete the REC task on the final scene. However, the process of navigating to the target object scene in most of above tasks merely consists of spatial movements without interaction with surrounding environments, such as opening closed objects or moving occlusion.

\begin{figure*}[htb]
    \centering
    \includegraphics[width=0.8\textwidth]{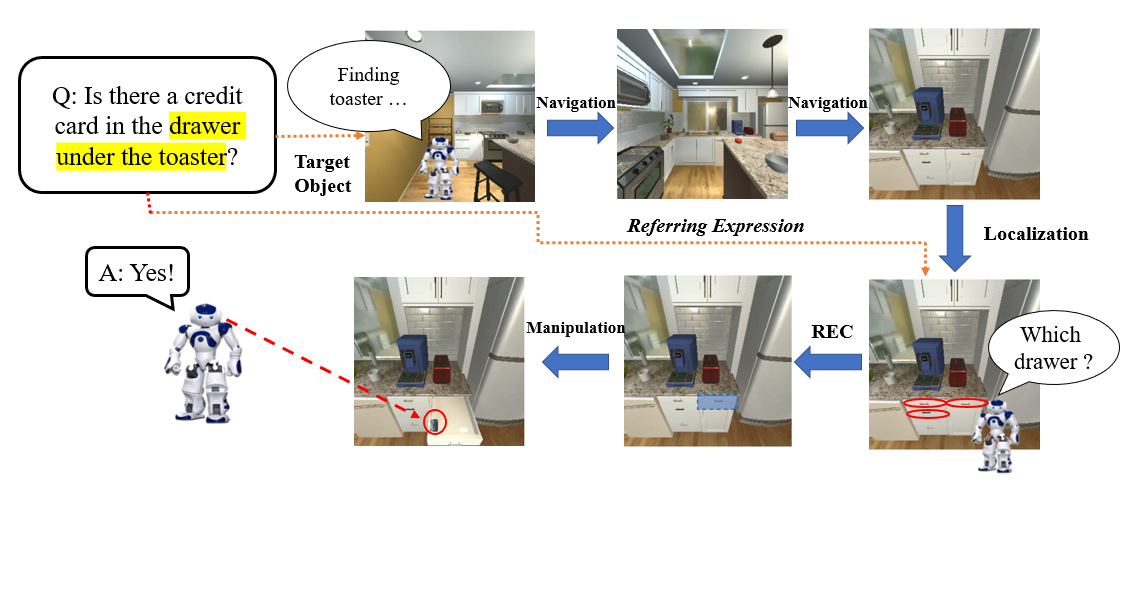}
    \caption{A demonstration of the Remote Embodied Manipulation Question Answering task. The agent needs to navigate to the goal position, localize the target object and perform manipulation to answer the question.}
    \label{fig:task}
\end{figure*}

Therefore, we introduce a novel embodied task \textbf{Remote Embodied Manipulation Question Answering (REMQA)} where the agent is required to navigate to a remote position and manipulate the target object, which can be precisely localized by referring expression comprehension. Then, the agent infers the answer to the question from the post-manipulation layout of objects. As we illustrate in Fig. \ref{fig:task}, the input question consists of a referring phrase explicitly referring to the target object (drawer). After navigating to the goal position (toaster), the agent needs to localize it by distinguishing it from other drawers with referring expression comprehension and conducting manipulation action (open drawer) to get the answer.

In this work, we focus on the referring expression comprehension problems for \textbf{Manipulation Question Answering (MQA)} task in a physically interactive environment. The main contributions of this work are listed below:
\begin{itemize}[itemsep=2pt,topsep=0pt,parsep=0pt]
\item \textbf{Problem.} a novel embodied robot task consists of vision perception, language comprehension and manipulation in an interactive environment, Remote Embodied Manipulation Question Answering.
\item \textbf{Dataset.} a benchmark dataset of proposed task with a set of indoor object arrangements of different rooms in an interactive environment and questions within referring expression about the objects in the environment.
\item \textbf{Method.} a framework to handle the proposed task in which Language Attention Network and 3D semantic memory prior-ed navigation are implemented. Experimental validation of the proposed model has been conducted in an interactive environment with the physical engine.
\end{itemize}
In the rest of the paper, Section \ref{sec:related work} presents a review of related works. Summary of data for pretraining and proposed benchmark dataset is introduced in Section \ref{sec:dataset}. Section \ref{sec:model} details our proposed model for the Remote Embodied Manipulation Question Answering task in an interactive environment. Section \ref{sec:exp} presents the experimental results and Section \ref{sec:future work} concludes this work. 
\section{Related Work}
\label{sec:related work}
\subsection{Referring Expression}
\textbf{Referring Expression on Static Dataset:} Most of works in referring expression comprehension focus on comprehension tasks in datasets built on classical static visual datasets (COCO, Flick et al.). Specifically, RE tasks can be categorized to two kinds with aspect to labels used for localization: 1) Referring Expression Comprehension (REC): a bounding box 2) Referring Expression Segmentation (RES): a segmentation mask. For REC task, Mao et al. \cite{mao2016generation}introduce the first CNN-LSTM method:MMI as a general solution to REC task. Yu et al. propose a visual comparative method (Visdif) to distinguish the target object from the surrounding objects rather than extracting features by CNN. Furthermore, Yu et al. \cite{yu2018mattnet} raise MAttNet: Modular Attention Network to decompose referring expressions into different modular channels for accurate matching. Besides CNN-LSTM methods, some works \cite{hu2017modeling,yang2019cross}present models of the relationship between images and expressions and some others\cite{li2021referring} utilize the pre-trained vision and language models for REC task. For RES task, Li et al.\cite{li2018referring} propose a multi-modal LSTM for vision and linguistic fusion. To obtain more accurate results for long referring expressions, Shi et al. \cite{shi2018key} employ an attention mechanism in raised keyword-aware network. Luo et al. \cite{2020Multi} introduce Multi-task Collaborative Network (MCN) as a joint learning framework of RES.

\textbf{Embodied Referring Expression:} Due to the absence of interaction in conventional referring expression tasks, researchers have recently tried to transplant referring expression tasks to embodied scenarios. Several ERE tasks and datasets have been released in recent years. Most proposed ERE tasks can be classified into two main categories with aspect to platform: 1) ERE task in manipulator scenario: INGRESS\cite{2020INGRESS} 2) ERE task in mobile navigation scenario: REVIERE\cite{qi2020reverie}, Touchdown-SDR\cite{2019TOUCHDOWN}, REVE-CE\cite{REVECE}, ALFRED\cite{2020ALFRED}
The community havs developed several methods that enable agents to tackle embodied tasks that require active interaction with the environment.
Wu et al.\cite{qi2020reverie} propose a Navigator-Pointer model as a baseline for REVIERE dataset. Gao et al. employ room object-aware attention mechanism and transformer architecture in REVIERE. Lin et al. \cite{lin2021scene} pre-train agent with cross-modal alignment sub-tasks for ERE task.
\subsection{Embodied Robot Task}
As an intersection of robotics, computer vision and natural language processing, the study of embodied robot tasks has gained much attention from all the above fields. A wide variety of embodied tasks has been formulated in recent years. The off-shelf embodied robot tasks can be categorized into two main types: Visual Navigation and Question Answering.

\textbf{Visual Navigation Task:} Visual Language Navigation(VLN)\cite{anderson2017visionandlanguage}, Visual Semantic Navigation (VSN)\cite{2014VSN} requires the agent to actively navigate to the goal position following linguistic information: language instructions for VLN  semantic labels for VSN. Anderson et al.\cite{anderson2017visionandlanguage,anderson2018evaluation} introduce the seq-to-seq framework and evaluation metrics for VLN task.

\textbf{Question Answering Task:}  Antol et al. \cite{2014VSN} firstly formulate  Visual Question Answering (VQA).  The agent needs to infer the answer from the image passively, which only relies on understanding questions and images. Many works\cite{das2017visual,le2020hierarchical} on VQA have been proposed in the past decade. In an EQA task\cite{EQA}, the agent is randomly spawned in a 3D environment and should explore the scenario with egocentric vision and answer the question with the final scene. Most of the methods proposed for EQA are based on Reinforcement Learning (RL)\cite{EQA,wijmans2019embodied}. Yu et al.\cite{yu2019multitarget} extend EQA to multi-target scenario. Tan et al. \cite{MultiEQA} employ a multi-agent system for EQA. Interactive Question Answering is an extension of EQA raised by Gordon et al.\cite{IQA}. In IQA, the agent needs to do some simple standard virtual interactions (\textbf{e.g.} open the fridge) with the environment. Deng et al. \cite{deng2020mqa} introduce Manipulation Question Answering (MQA), where a fixed-base manipulator is required to manipulate objects in the cluttered scene to render more information about objects initially unseen and answer the question better.

 In following Table \ref{tab:cmp}, we compare the difference of several robot embodied tasks mentioned above. As we can see, only our proposed task has taken all 4 mentioned modules into consideration.
\begin{table}[htbp]
	\centering
	\begin{tabular}{|c|c|c|c|c|c|c|}
		\hline
		&VSN/VLN& VQA& EQA & IQA & MQA & Ours \\ \hline
		Language & $\checkmark$  & $\checkmark$ & $\checkmark$ & $\checkmark$ &  $\checkmark$&  $\checkmark$ \\ \hline
		Navigation & $\checkmark$ & - & $\checkmark$ & $\checkmark$ & -& $\checkmark$\\ \hline
		Interaction & - & - & - & $\checkmark$  & $\checkmark$ &  $\checkmark$       \\ \hline
		Manipulation & - & - & - & - & $\checkmark$ &  $\checkmark$        \\ \hline
	\end{tabular}
	\\
	\caption{Comparison of different embodied robot tasks}
	\label{tab:cmp}
\end{table}
\section{Dataset}
\label{sec:dataset}
\subsection{Embodied Environment}
Training and evaluating an interactive agent in a real environment is temporarily uneconomic considering costs, time and generalizability. Therefore, we adopt AI2-THOR, a photo-realistic 3D environment simulator designed for embodied AI research\cite{kolve2017ai2}, as our learning framework. The AI2Thor simulator consists of 120 different room layouts of 4 categories (Bedroom, Living room, Kitchen, and Bathroom), with 30 layouts for each category. We choose an extension of the AI2-THOR simulator: ManipulaTHOR, which has the same scenes as AI2-THOR and a realistic Kinova 6-DOF arm added to the agent\cite{ehsani2021manipulathor}. ManipulaTHOR allows agents to interact physically with objects at a low control level via arm manipulators. Besides, several sensors including RGB-D frame, agent location, and arm configuration at the arm-joint level are also available, enabling us to render the metadata of agent to design embodied tasks.

\subsection{Pretraining Data}
\label{sec:pretrain}
We build our datasets for pretraining our instance segmentation and referring expression comprehension modules. The agent samples more than 10000 images from all scenes in AI2-THOR with the BFS strategy. Annotations including semantic labels, positions in the scene and ground-truth segmentation masks are automatically generated from the metadata of the simulator.

\begin{figure*}[htbp]
    \centering
    \includegraphics[width=0.9\textwidth]{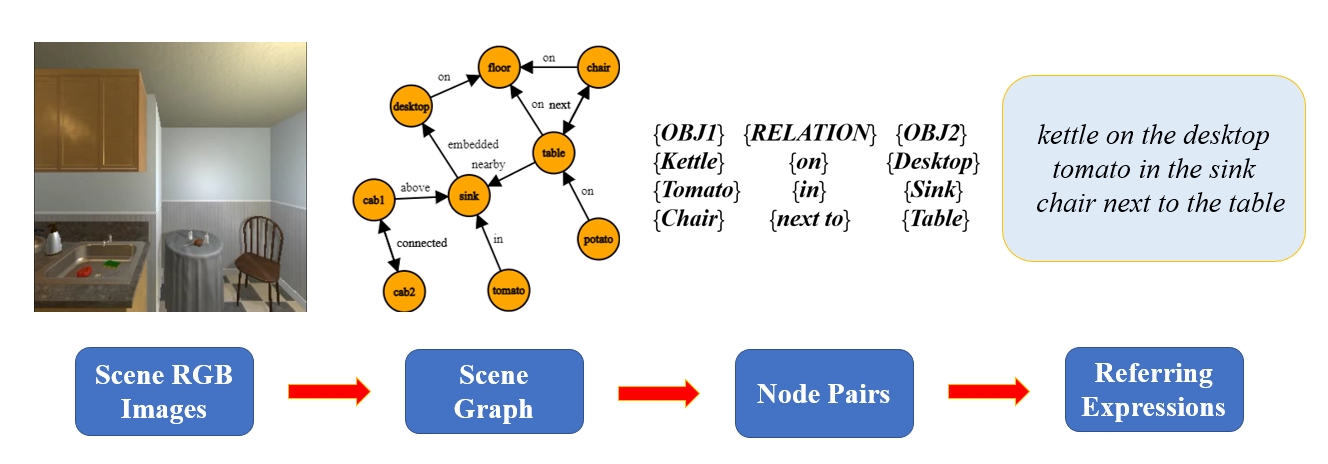}
    \caption{An overview of our referring expression data generation pipeline}
    \label{fig:scene-data}
\end{figure*}

We build the corresponding semantic scene graph incrementally from metadata during the sampling for referring expressions used for REC module pretraining. As shown in Fig. \ref{fig:scene-data}, scene frames and metadata of objects seen in the frame are sampled during navigation. By updating frames during sampling, we add new objects as nodes and spatial relationships between objects as edges into scene graph. Two metrics assign the spatial relationship between two objects: distance between central points $l$ and 3D IoU(intersection over union ) of ground-truth bounding boxes $S_{IoU}$. 
Then, we build our referring expression set with node-edge-node pairs in scene graph in form of  template.$RE: \quad   \textit{the \quad \{OBJ1\} \quad \{RELATION\} \quad the \quad \{OBJ2\} }$, where $\textit{ \{OBJ1\},\{OBJ2\}}$ represent objects and  $\textit{\{RELATION\}}$ represents spatial relationship between them.
\subsection{Summary of REMQA dataset}
 We select 60 types of most frequently seen objects out of objects in AI2-THOR to build our REMQA dataset. Similar to the construction of dataset for REC module pretraining, we generate our questions using node-edge-node pairs from the scene graph. We designed three kinds of questions in our REMQA dataset: COUNTING, EXISTENCE, and SPATIAL questions. The templates of 3 kinds of questions are shown in Table. \ref{tab:questions}
 \begin{table}[h]
	\caption{Question Templates}
	\centering
	\begin{tabular}{|c|c|}
		\hline
		EXISTENCE      &  \textit{Is there a \{OBJ1\} in the  \{\textbf{RE}(OBJ2)\}?}\\\hline
		COUNTING        &  \textit{How many \{OBJ1\} are there in the \{\textbf{RE}(OBJ2)\}?}\\\hline
		SPATIAL  &  \textit{Is there a \{OBJ1\} \{RELATION\} the \{OBJ2\}?}\\\hline
	\end{tabular}
	\label{tab:questions}
\end{table}

As the statistics of our proposed REMQA dataset shown in following Table. \ref{tab:dataset_split} , REMQA dataset is composed of 120 scenes. Among the dataset, we use 100 scene series for training and 20 for testing. There are altogether 4072 questions of three kinds.

\begin{table}[h]
	\centering
	\caption{REMQA dataset split}
	\begin{tabular}{|c|c|c|c|c|}
		\hline
		   & Scenes & EXISTENCE & COUNTING & SPATIAL \\
		\hline
		train & 100 & 1653 & 647 & 1083  \\ 
		\hline
		test & 20 & 324 & 143 & 222  \\ 
		\hline
		all & 120 & 1977 & 790 & 1305 \\
		\hline
		avg. Length & - & 9.5 & 11.8 & 7.1 \\
		\hline
	\end{tabular}
	\label{tab:dataset_split}
\end{table}
Similar to off-shelf EQA datasets\cite{EQA,IQA}, each scene is associated with multiple scene configurations that result in different answers to the same question. For every task sample in the dataset, a question with a referring phrase, ground-truth answer, scene configuration, the initial scene of the target object and final scene after manipulation are included.
\section{Proposed Approach}
\label{sec:model}
\begin{figure*}[htbp]
    \centering
    \includegraphics[width=0.9\linewidth]{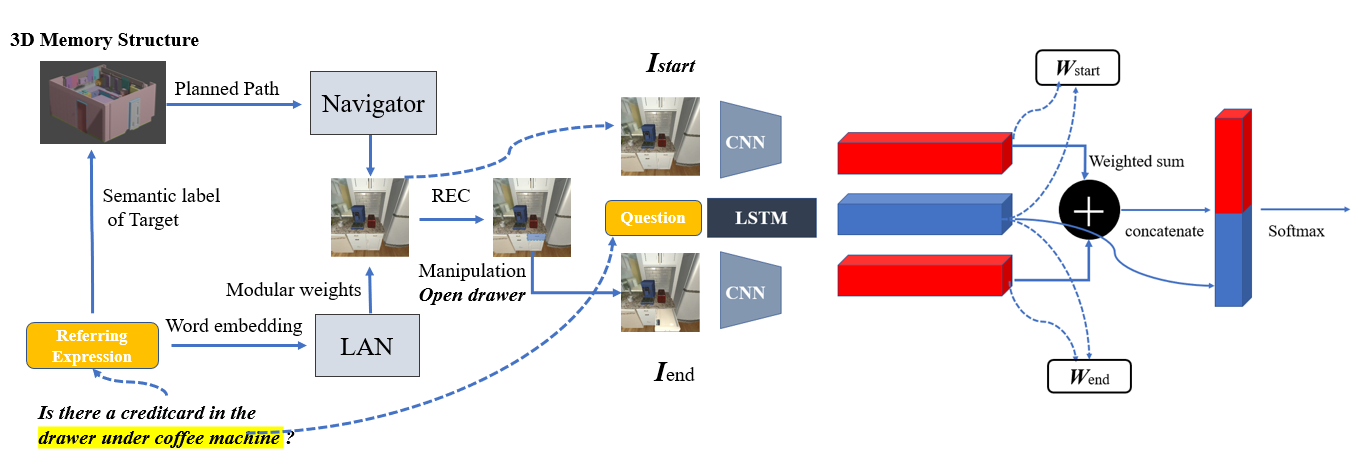}
    \caption{The overall architecture of proposed framework: The referring expression in input question passes through LAN for REC. After the manipulation, the frames before and after manipulation along with embedded question are fused by CNN-LSTM QA part }
    \label{fig:model}
\end{figure*}

As shown in Fig. \ref{fig:model},we have proposed a general framework to for REMQA task which consists of three main parts: 1) Navigation module 2) Referring expression comprehension module 3) Manipulation Question Answering module.
\subsection{Navigation}
\begin{figure*}[htb]
    \centering
    \begin{subfigure}{0.3\textwidth}
        \includegraphics[scale=0.3]{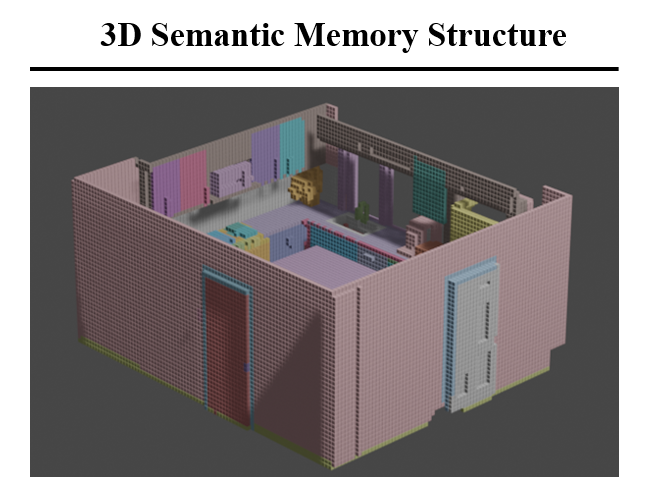}
        \caption{The illustration of 3D semantic memory structure consist of semantic voxels}
        \label{fig:map1}
    \end{subfigure}
    \hfill
     \hspace{0.2in} 
     \centering
    \begin{subfigure}{0.65\textwidth}
        \centering
        \includegraphics[scale=0.31]{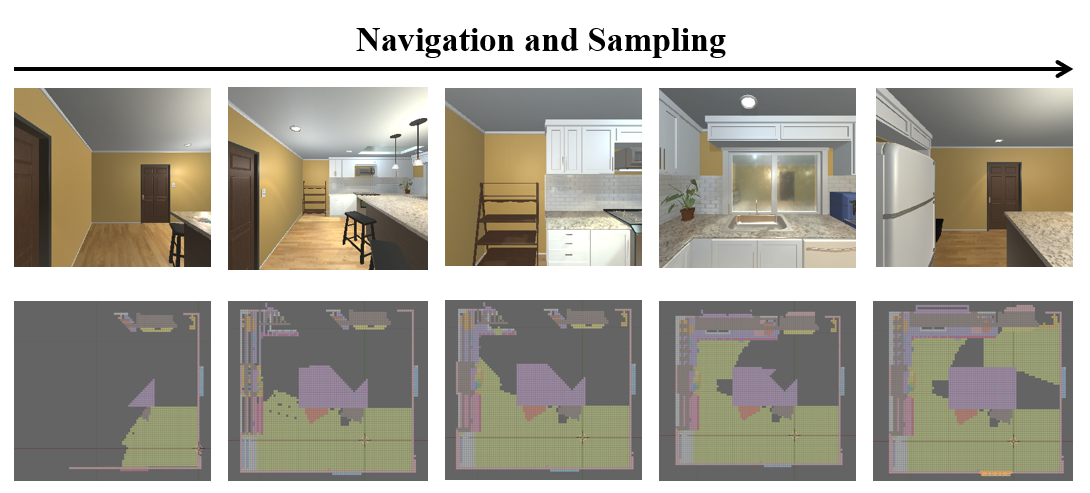}
        \caption{The RGB frames of the agent and incremental semantic map during sampling. The room is a kitchen scenario provided in AI2THOR simulator}
        \label{fig:map2}
    \end{subfigure}
    \label{fig:map}
\end{figure*}

To enhance the performance of the navigation module with information of task scenarios, we build a knowledge-prior visual semantic navigation model based on a scene graph and semantic map of a given indoor environment. The construction of the scene graph is mentioned in Section \ref{sec:pretrain}. To construct the semantic map, the agent first navigates in AI2THOR scenes with pre-designed sampling paths to incrementally build the 3D semantic memory structure of every room layout from metadata. The illustration of constructed structure is presented in Fig.\ref{fig:map1}. Every voxel and its color in the structure represents a tiny cubic space occupied by an object and the corresponding object type.
Meanwhile, the 3D memory with semantic labels is dynamically transformed into a 2D semantic map by dimensional reduction during the sampling. As shown in Fig.\ref{fig:map2}, the agent samples the RGB frames and metadata of environment when navigating along the given path in each room of AI2THOR. The semantic map of the task scene is updated at every time step.

Employing generated 2D semantic maps as prior knowledge, the agent first locates the target object's position by searching its semantic label in the map. Then the shortest path between the initial agent position and the target object is planned by the Floyd algorithm in the semantic map for the navigation task.

\subsection{Referring Expression Comprehension}
We build out referring expression comprehension module by adopting Language Attention Network (LAN) a modular design from MAttNet\cite{yu2018mattnet}. The LAN decomposes referring expressions in the form of word embedding into three modular components: subject attributes, location and spatial relationship to other objects. For every module, a phrase embedding is provided to calculate the matching scores of corresponding area in the given image without affecting each other. The overall matching scores weighted by modular weight are calculated to match objects with expressions.

\subsection{Manipulation Question Answering}
We implement a Manipulation Question Answering (QA) module based on LSTM network. The upstream REC module pass the semantic label of target object which is usually an occlusion (e.g. Fridge, Cabinet) to a classifier. The classifier will choose the manipulation type from the action set: $\mathcal{A}=\{\textit{Open},\textit{Move},\textit{Pickup}\}$ according to the type of target object (e.g. Open Fridge, Move chair,Pick up book). Then the agent moves its arm until the arm end reaches the vicinity of the target object by rendering the position from 3D semantic memory. Due to the incompleteness of the dynamic simulation of Manipulathor, the manipulation will be automatically conducted to objects within end sphere. Hence, our proposed manipulation module has no action generation and dynamic planning.

 We denote the initial RGB frame before manipulations as $I_{start}$ and the frame after manipulations as $I_{end}$. 
The answering module encodes $I_{start},I_{end}$ frames with CNN and the input question with a 2-layer LSTM network. Then, attention weights based on image-question similarity are computed to fuse the features of two images and 
image features with an encoded question. A softmax classifier outputs predictions from the space of possible answers with the fused feature passed through.
\begin{table}[h]
    \centering
	\caption{Comparison of VSN methods}
	\begin{tabular}{ccccc}
		\toprule
		\multicolumn{1}{c}{ Methods}&
		\multicolumn{2}{c}{Max steps=25}&
		\multicolumn{2}{c}{Max steps=50}\cr
    	\cmidrule(lr){2-3} \cmidrule(lr){4-5}
		&Success&SPL&Success&SPL\cr
		\midrule
	    Random  &0.013&0.006 &0.035&0.013\cr
		A3C     &0.210&0.129 &0.241&0.152\cr
		SAVN    &0.283&0.121&0.396&0.178\cr
		Scene Priors &0.264&0.117&0.376&0.164\cr
		Ours    &\textbf{0.566}&\textbf{0.461} &\textbf{0.729}&\textbf{0.595}\cr
		\bottomrule
	\end{tabular}
	\label{tab:nav_results}
\end{table}
\begin{table}[h]
    \centering
	\caption{Comparison of REC methods}
	\begin{tabular}{ccc}
		\toprule
		\multicolumn{1}{c}{Methods}&
		\multicolumn{2}{c}{Pred@0.5}\cr
		\cmidrule(lr){2-3}
		&Val&Test\cr
		\midrule
	    MCN  &71.04&73.16 \cr
	    CGAN &73.18&76.94 \cr
		BiLSTM\cite{hu2017modeling} + detectron2 &66.57&70.45 \cr
		LAN+MaskRCNN & 70.27 & 74.12 \cr 
		LAN+detectron2(Ours) &\textbf{75.20}&\textbf{77.52} \cr
		\bottomrule
	\end{tabular}
	\label{tab:rec_results}
\end{table}
\begin{table*}[h]
	\centering
	\caption{Evaluations of QA models on REMQA dataset}
	\begin{tabular}{cccccccccc}
		\toprule
		\multicolumn{1}{c}{ Methods}&
		\multicolumn{3}{c}{EXISTENCE}&
		\multicolumn{3}{c}{COUNTING}&
		\multicolumn{3}{c}{SPATIAL}\cr
     	\cmidrule(lr){2-4} \cmidrule(lr){5-7} \cmidrule(lr){8-10}
		&$S_{N}$&$S_{L}$&$S_{QA}$&$S_{N}$&$S_{L}$&$S_{QA}$&$S_{N}$&$S_{L}$&$S_{QA}$\cr
		\midrule
	    EQA  &0.509&0.389&0.330
	    &0.455&0.350&0.217
	    &0.468&0.396&0.131\cr
		IQA &0.642&0.596&0.488
		&\textbf{0.769}&\textbf{0.692}&0.510
		&0.653&0.550&0.216\cr
		Ours &\textbf{0.738}&\textbf{0.620}&\textbf{0.540}
		&0.671&0.608&\textbf{0.559}
		&\textbf{0.752}&\textbf{0.595}&\textbf{0.284}\cr
		\bottomrule
	\end{tabular}
	\label{tab:remqa_result}
\end{table*}

\begin{figure*}[h]
    \centering
    \includegraphics[width=0.8\textwidth]{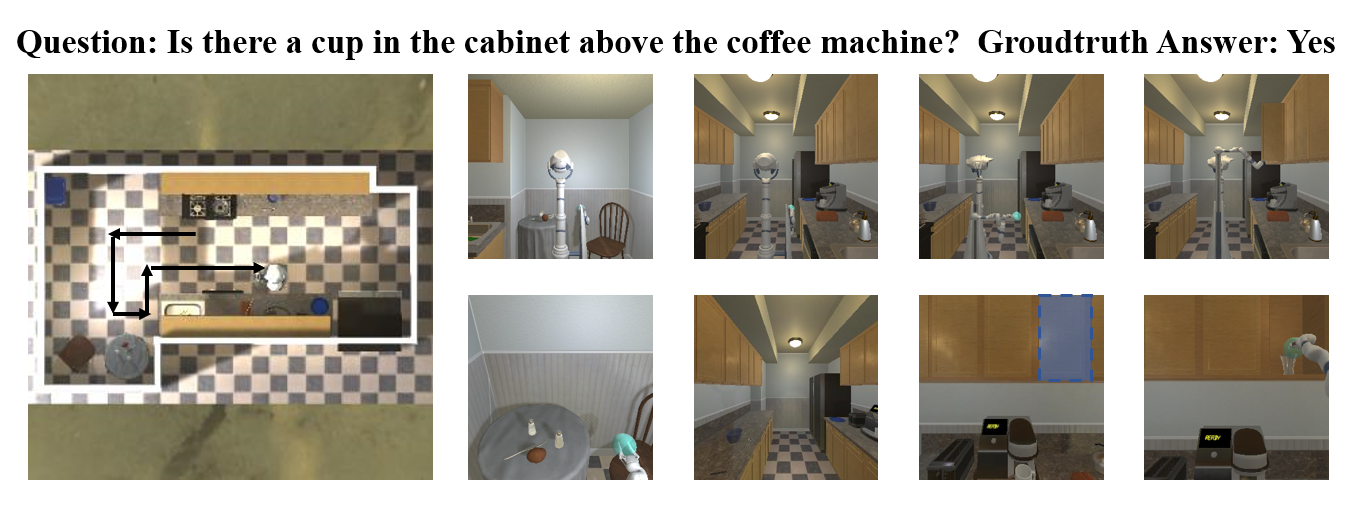}
    \caption{A qualitative example with the question. The trajectory of agent is presented on the left side and some rendered images of the agent egocentric view and third party camera on the right  }
    \label{fig:qualitative}
\end{figure*}
\section{Experiments}
\label{sec:exp}
In this section, we first introduce the experimental settings and evaluation metrics. Then, we present the analysis of the quantitative results of our proposed method and several baseline variants. Finally, an MQA task sample is selected to illustrate how the agent navigates, comprehends and answers the question as qualitative results.
\subsection{Embodied Referring Expression}
\textbf{Experimental settings:} To comprehensively illustrate the performance of our model on embodied referring expression task, we analyze the ERE results in two stages: the performance on VSN and the performance on REC toward ground-truth final scenes assuming that agent navigates successfully in the first stage.
Two kinds of metrics are raised for evaluation of VSN task: the success rate of navigation $s$ and Success rate weighted by
Path Length (SPL)\cite{anderson2018evaluation}. The SPL calculates the success rate of the VSN task weighted by the ratio of shortest path distance from the starting position to the goal by the path distance the agent actually takes. For REC performance, we adopt prec@X\cite{2020Multi} measures the percentage of test images with an intersection over union (IoU) score higher than the threshold $X$ and here we set $X=0.5$.

In the VSN stage, we compared our proposed semantic map priored navigation model with Random agent, traditional RL methods: A3C and other embodied navigation methods that use metadata of scene in AI2THOR as priors for navigation: SAVN\cite{2019SAVN}, Scene Priors\cite{prior}. All baselines are evaluated on seen objects and known objects. Results are summarized in following Table \ref{tab:nav_results}.Taking advantage of the 2D semantic map obtained from the 3D memory structural, our proposed navigation model outperforms all baseline methods by a large margin in terms of both success rate and SPL metrics. It is noted that the 2D semantic map only represents information of scenes seen before. Our proposed method can be improved in the generalization of novel scenes.

In the REC stage, a set of off-shelf models proposed for referring expression tasks are selected as baselines: MCN\cite{2020Multi}, CGAN\cite{luo2020cascade} and combinations of language parser and object detector with other backbones. The validation and test REC datasets are split from referring phrase set in Section \ref{sec:pretrain}. Results are summarized in following Table \ref{tab:rec_results}. Our proposed REC model outperforms all baseline models on both validation and test set. It is worth noting that there is no significant margin between our model and SoTA baseline models. However, the performance gap between two recombined baselines is much larger than SoTAs. We infer the reason that SoTAs and our model all adopt the modular module to process the referring expressions into multi-modal channels. The result can also validate the effectiveness of the modular module that BiLSTM + detectron2  baseline performs much lower than LAN+MaskRCNN since Bi-LSTM can only encode the referring expressions without modular extraction.

\subsection{MQA on REMQA dataset}
To validate the effectiveness of our proposed model, we separately compare our navigation and comprehension results with the state-of-the-art methods (SoTAs) on our proposed benchmark dataset. The results are presented in Table. \ref{tab:remqa_result}. The $S_{N}, S_{L}$ denotes the ratio of samples successfully navigated to the goal position and successfully localized at the goal position with REC. The $S_{QA}$ represents the rate of correctly answered questions out of all samples in the REMQA dataset.

The results show that our framework for the REMQA task outperforms EQA and IQA models in most metrics except $S_N$ and $S_L$ in COUNTING questions. Most of the objects mentioned in COUNTING questions are receptacles (e.g.cabinets, drawers) that are many individuals in the same scene. IQA model is trained on a larger dataset (IQUAD v1 with 75000 questions) and may better distinguish instances from the same type. We also can notice that EQA model performs far below IQA and our model due to the absence of the ability to interact with the environment. Therefore, the effectiveness of manipulations for the REMQA task can be validated.

\subsection{Qualitative Results}
To illustrate how our agent navigates and manipulate in proposed Embodied Environments, we select a task sample from proposed REMQA dataset as a qualitative example. As shown in following Fig. \ref{fig:qualitative}, the agent actively explores in kitchen scenarios to find the coffee machine mentioned in the input question. When successfully navigated, agent localizes the target cabinet with REC and move its arm to open it to find whether there is a cup for final answer.

\section{Conclusion}
\label{sec:future work}
In this work, We have proposed a brand new embodied robot task Remote Embodied Manipulation Question Answering (REMQA) in a physically interactive environment. We build a benchmark dataset for the task by scene graph generation. To solve this problem, we propose a general framework consisting of a VSN module with scene semantic map as priors, a LAN for referring expression comprehension and manipulation decision, and a CNN-LSTM network for question answering. The experimental results on the new benchmark dataset validate the effectiveness of our proposed model. This task still  challenges to our proposed method in more complicated question and multi-stage task. For the future study, a validation experiment in real robot platforms and physical environment is expected to validate the ability of to conduct more complex actions.

\clearpage


\bibliographystyle{IEEEtran}
\bibliography{refs}

\end{document}